\documentclass[10pt,twocolumn,letterpaper]{article}

\usepackage{colortbl}
\usepackage{amsfonts}
\usepackage{algorithmic}

\usepackage{lineno}
\usepackage{multirow}
\usepackage{pifont}
\usepackage{graphicx}
\usepackage[dvipsnames]{xcolor}
\usepackage{amsmath}
\usepackage{amssymb}
\usepackage{booktabs}
\usepackage{overpic}
\usepackage{float}
\usepackage[linesnumbered, boxed, ruled]{algorithm2e}
\usepackage{array}
\usepackage{soul}
\usepackage{makecell}
\usepackage[pagenumbers]{cvpr} 

\usepackage{tcolorbox}
\usepackage{balance}

\definecolor{boost}{RGB}{243,220,205}
\definecolor{adapt}{RGB}{232,243,225}
\definecolor{boost_line}{RGB}{197,90,17}
\definecolor{adapt_line}{RGB}{84,130,53} 

\definecolor{tablecolor}{gray}{.93}
\definecolor{decline_color}{RGB}{75,140,174}

\definecolor{cvprblue}{rgb}{0.21,0.49,0.74}
\usepackage[pagebackref,breaklinks,colorlinks,allcolors=cvprblue]{hyperref}
\definecolor{lzh}{RGB}{255, 192, 203}
\definecolor{lmx}{RGB}{197, 90, 17}

\usepackage[symbol]{footmisc}

\def\paperID{***} 
\def\confName{ICCV}
\def\confYear{2025}

\title{Beyond Flat Text: Dual Self-inherited Guidance for Visual Text Generation}
\newcommand*{\affaddr}[1]{#1} 
\newcommand*{\affmark}[1][*]{\textsuperscript{#1}}

\author{%
Minxing Luo\footnote{}~~\affmark[1], Zixun Xia\footref{note1}~~\affmark[1], Liaojun Chen\affmark[1], Zhenhang Li\affmark[2], Weichao Zeng\affmark[2], Jianye Wang\affmark[1],\\ 
Wentao Cheng\affmark[1], Yaxing Wang\affmark[1], Yu Zhou\affmark[1] and Jian Yang\affmark[1] \\
\small{\affaddr{\affmark[1] VCIP, CS, Nankai University,~~}}\small{\affaddr{\affmark[2] Institute of Information Engineering, Chinese Academy of Sciences}}\
}

\begin{document}
\twocolumn[{%
\renewcommand\twocolumn[1][]{#1}%
\maketitle
\includegraphics[width=1.0\linewidth]{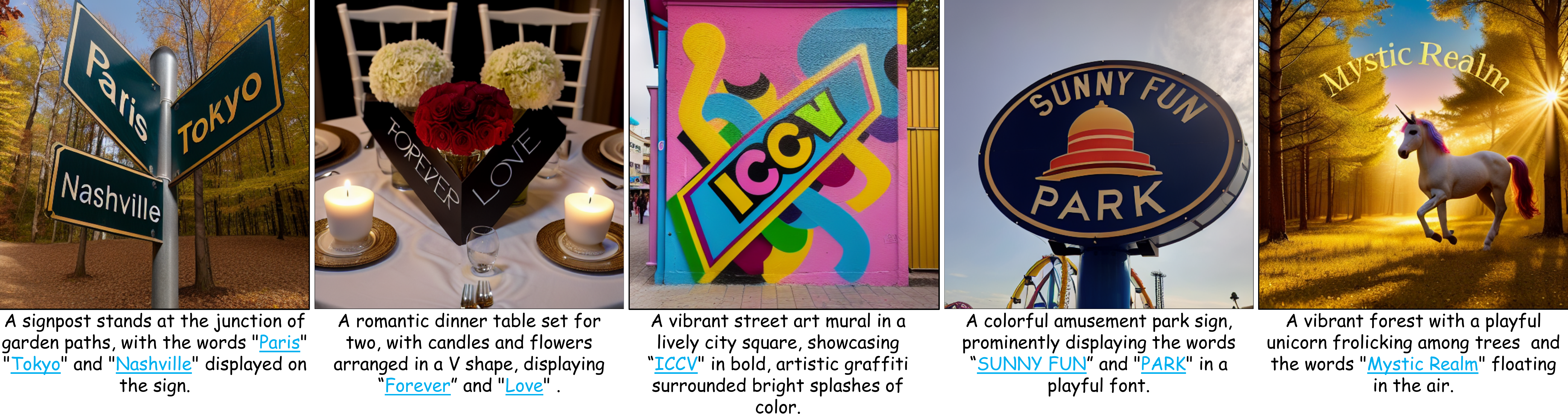}
\vspace{-20pt}
\centering
\captionof{figure}{\textbf{STGen for visual text generation in challenging layout.} 
\textit{Using a pre-trained visual text generation model (e.g., AnyText~\cite{tuo2023anytext}), our method, STGen, guides the model to adjust the text region in latent space during image synthesis, producing images that more faithfully represent the input prompt with precise visual text.} 
\vspace{1em}
}
\label{fig:teaser}
}]
\footnotetext[1]{\label{note1} Equal contribution.}
\begin{abstract}
In real-world images, slanted or curved texts, especially those on cans, banners, or badges, appear as frequently, if not more so, than flat texts due to artistic design or layout constraints. While high-quality visual text generation has become available with the advanced generative capabilities of diffusion models, these models often produce distorted text and inharmonious text backgrounds when given slanted or curved text layouts due to training data limitations. In this paper, we introduce a new framework, STGen, which accurately generates visual texts in challenging scenarios (\eg, slanted or curved text layouts) while harmonizing them with the text background. Our framework decomposes the visual text generation process into two branches: (i) \textbf{Semantic Rectification Branch}, which leverages the ability in generating flat but accurate visual texts of the model to guide the generation of challenging scenarios. The generated latent of flat text is abundant in accurate semantic information related both to the text itself and its background. By incorporating this, we rectify the semantic information of the texts and harmonize the integration of the text with its background in complex layouts. (ii) \textbf{Structure Injection Branch}, which reinforces the visual text structure during inference. We incorporate the latent information of the glyph image, rich in glyph structure, as a new condition to further strengthen the text structure. To enhance image harmony, we also apply an effective combination method to merge the priors, providing a solid foundation for generation. Extensive experiments across a variety of visual text layouts demonstrate that our framework achieves superior accuracy and outstanding quality. The code will be available at \href{https://github.com/Tony-Lowe/STGen}{https://github.com/Tony-Lowe/STGen}.
\end{abstract}

\section{Introduction}
\label{sec:intro}
Visual text generation is an emerging yet challenging research area in image generation because text is fine-grained and difficult to balance with the image. The current methods~\cite{srnet, yang2020swaptext, qu2023exploring,krishnan2023textstylebrush,subramanian2021strive,zhu2024visualtextgenerationwild,chen2024textdiffuser,saharia2022photorealistic,balaji2022eDiff-I,liu2023character,ma2023glyphdraw,yang2024glyphcontrol,liu2024glyph,tuo2023anytext,DBLP:conf/eccv/ChenHLCCW24} proposed a series techniques to address it.
Among them, diffusion based methods (AnyText~\cite{tuo2023anytext}, GlyphControl~\cite{yang2024glyphcontrol}, DiffText~\cite{zhang2024brush} \etc) can create new images with integrated text, raising an inevitable challenge to achieve both accurate visual texts and harmonious image content. Notably, AnyText~\cite{tuo2023anytext} distinguishes itself by producing impressive images integrated with outstanding multilingual text. It opens an era of universal visual text generation using large pretrained Visual Text Generation Model (VTGM). 

\begin{figure}[t]
    \centering
    \includegraphics[width=1.0\linewidth]{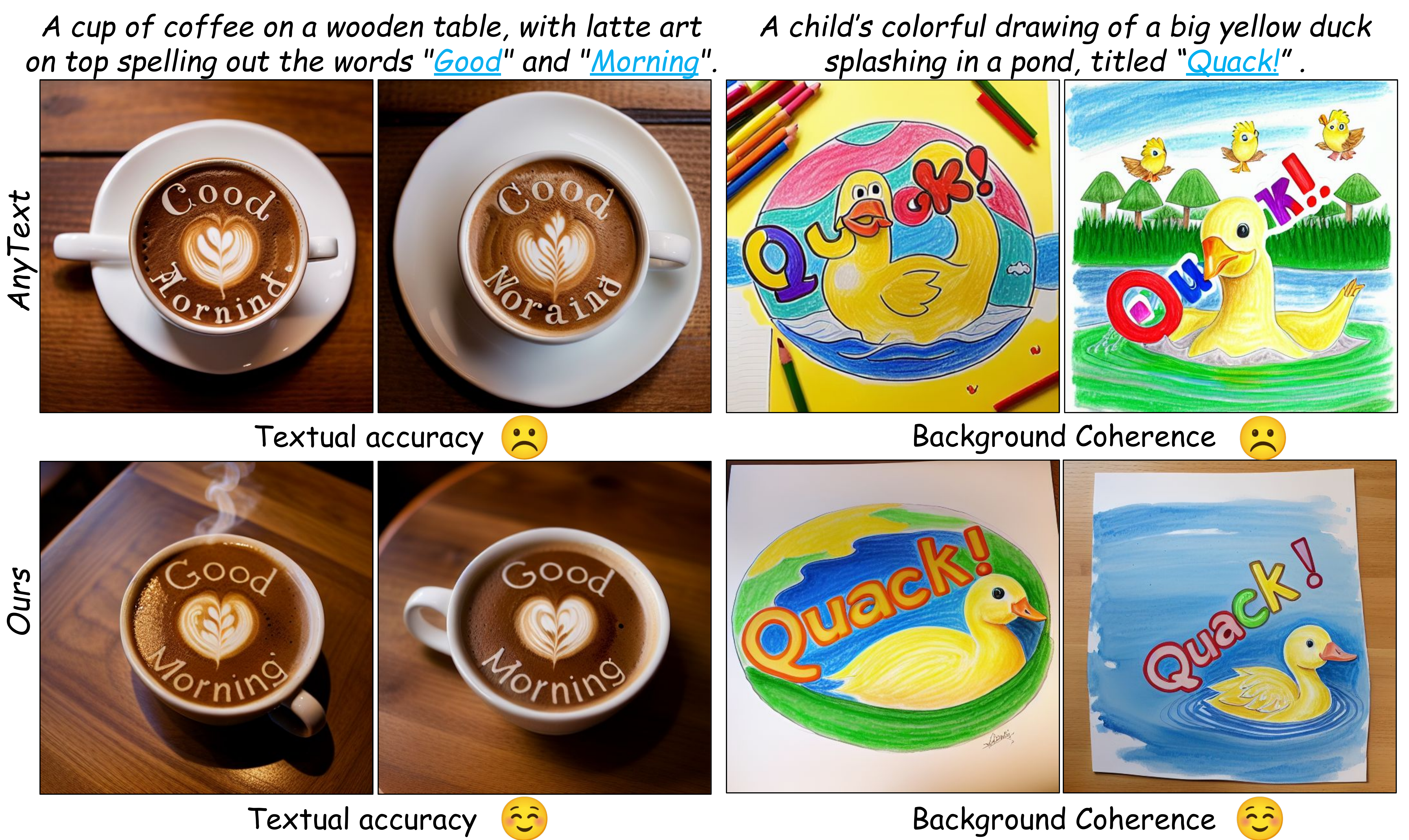}
    \vspace{-18pt}
    \caption{\textbf{Failure cases of AnyText~\cite{tuo2023anytext}}. \textit{The top row illustrates two failure cases: textual distortion (left) and background occlusion (right). The bottom row displays results using our method.
    }}
    \vspace{-15pt}
    \label{fig:2problems}   
\end{figure}

\begin{figure}[t]
    \centering
    \includegraphics[width=1.0\linewidth]{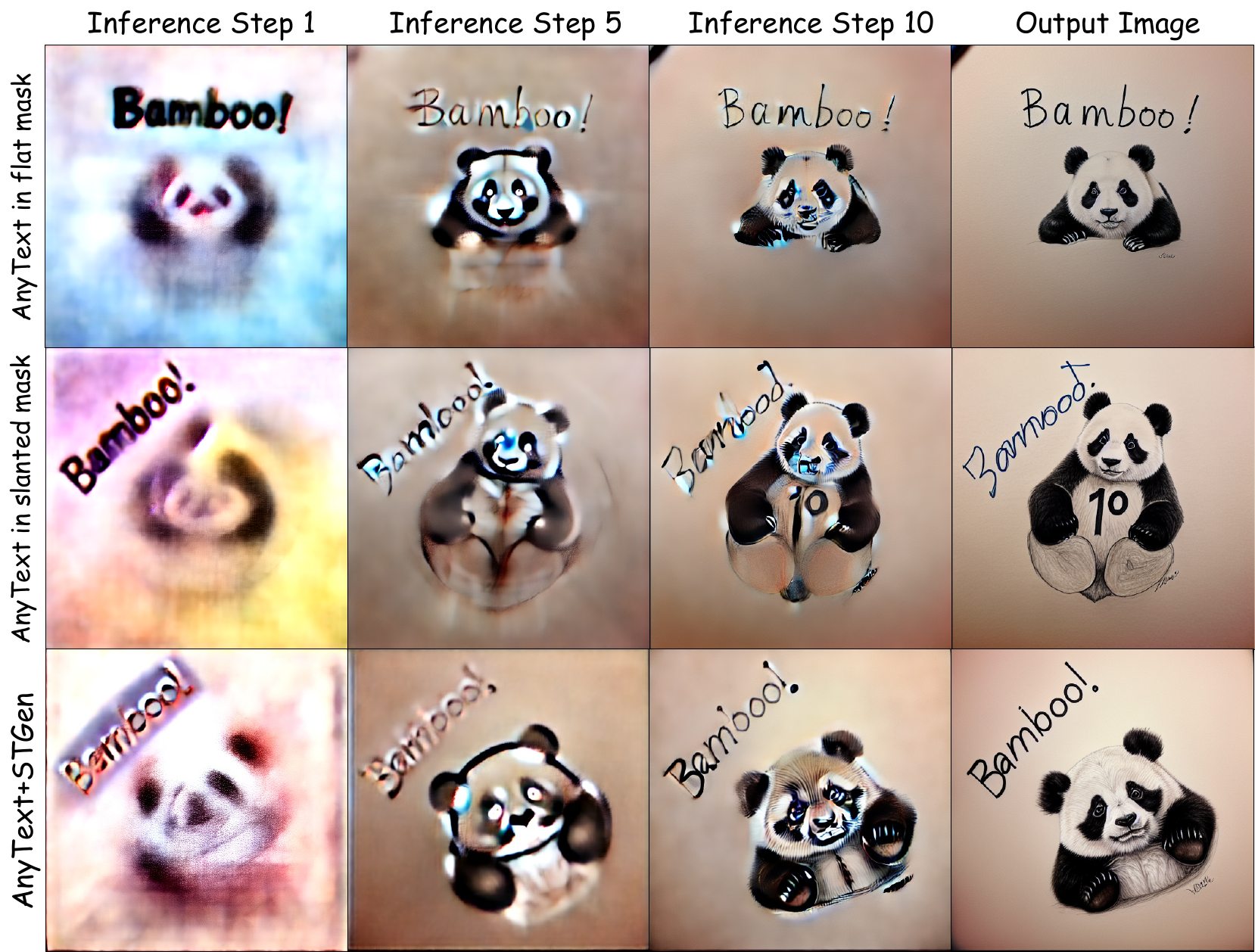}
    \vspace{-10pt}
    \caption{\textbf{Comparison of predicted $x_0$ under different inference steps.} \textit{The first and second rows show AnyText’s intermediate predicted $x_0$ for flat and slanted masks, respectively. 
    While predictions remain stable under flat masks, the $x_0$ prediction drifts during inference with slanted masks. Our method effectively guides the model to maintain accuracy in visual texts.}}
    \vspace{-10pt}
    \label{fig:demo_inter}
\end{figure}

Although large pretrained VGTMs take a significant step towards universal visual text generation, they still struggle to handle users' diverse inputs, such as commonly seen slanted or curved texts in real-world images. Given these user inputs, the model often leads to text distortion and background occlusion, as shown in \cref{fig:2problems}. This is because the latent space gradually becomes blurry and distorted (as shown in \cref{fig:demo_inter}) due to insufficient data in such scenarios, thus the model cannot effectively maintain the structure and semantic information in the text region as it does when processing a flat mask.




One naive solution is to train VTGM on a more diversified dataset that covers various text configurations. But it is resource-intensive and data distribution across multiple text layouts cannot be guaranteed. Even if they are trained on those data, the results may not be satisfying \cite{samuel2024norm,haviv2024not}.

To overcome this shortcoming, we propose a plug-and-play method named Slanted Text Generation (STGen), which corrects text regions in latent space during inference. Specifically, our approach employs a dual-branch framework. The first branch, the \textit{Semantic Rectification Branch (SRB)}, utilizes a latent generated using the same prompt, but with a simplified shape, as a robust semantic prior. This branch simultaneously rectifies distorted text predictions and harmonizes the text with its background. For complex visual text generation, such as text layouts composed of multiple tilted or curved sections, we propose a \textit{Divide and Conquer} strategy to efficiently reconfigure the text shape and obtain a reasonable semantic prior. 

The second branch, \textit{Structure Injection Branch (SIB)}, extracts rich structural information from glyphs and injects it into the latent space as a structural prior, further enhancing the accuracy of the visual text. Rather than simply merging the two priors, we adopt a novel combination method for optimized integration for better coherence in latent space. Together, the dual-branch framework offers effective guidance for the generation process without requiring additional training. 

To the best of our knowledge, our method is the first tailored specifically for generating visual texts in complex layouts and achieves state-of-the-art results as shown in \cref{fig:teaser,fig:qualitative,fig:bc} without additional training. Our framework effectively improves visual text accuracy across various approaches, including early-stage GlyphControl, DiffText, which receives no specified training for visual text generation, and AnyText. We conduct thorough experiments to demonstrate the superiority of our method and the effectiveness of each component in generating complex visual texts while enhancing overall image quality and coherence. The main contributions of our work are summarized as follows:
\begin{enumerate}[label=(\roman*)]
\item We present a new challenge: generating visual text in complex layouts with diffusion models. To tackle this, we introduce STGen, a dual-branch, training-free framework that can be easily integrated with existing text generation models. This framework enables the model to generate text in complex layouts by incorporating both structural and semantic priors.
\item  We introduce an effective method to combine given priors within the latent space. This approach seamlessly merges priors and the latent, ensuring a consistent latent range and resulting in a balanced, high-quality image. 
\item We propose a benchmark extending the AnyText-benchmark to evaluate texts in complex layouts, enabling comprehensive assessment across varying text slant difficulties.
\end{enumerate}



\section{Related Works}
\label{sec:rw}


Denoising diffusion probabilistic models (DDPMs)~\cite{ho2020denoising,ramesh2021zero,songscore,dhariwal2021diffusion,nichol2021improved,saharia2022photorealistic,ramesh2022hierarchical,rombach2022high,chang2023muse} have made significant contribution to the field of text-to-image synthesis. However, achieving precise control over image details remains challenging, particularly in visual text generation. 

 \subsection{Local Visual Text Blending}
 
Local Visual Text Blending methods synthesize localized images with single-line textual content. Existing approaches typically necessitate a background image for text insertion or modification. SynText~\cite{gupta2016synthetic} identifies image regions suitable for text placement and renders text accordingly. Following their work, SceneVTG~\cite{zhu2024visualtextgenerationwild} introduces a model to predict the text-generating area and specifically designs a framework that takes line masks and word masks separately to help generate texts in different shapes. SceneVTG is less user-friendly because it requires an erased image and lacks support for multilingual text generation. AnyTrans~\cite{qian2024anytrans} detects, removes, and replaces text with translations in the same region. The above methods require an input image to render or generate, which is good for editing, but cannot generate the whole image from scratch.
 
 \subsection{Global Visual Text Generation }

 Recent studies, such as Imagen~\cite{saharia2022photorealistic}, demonstrate that replacing CLIP text encoder~\cite{radford2021learning} with more advanced models like T5~\cite{raffel2020exploring} enhances visual text generation. Liu \etal~\cite{liu2023character} further replace the character-blind text encoder with a character-aware text encoder. GlyphByT5 series~\cite{liu2024glyph,liu2024glyph2} employ character-aware ByT5 encoder~\cite{liu2022character} and a new cross attention mechanism to compute the text region and image region separately. Although replacing text encoders appears straightforward, it still struggles to generate complex characters such as Korean, Japanese, and Chinese. GlyphDraw~\cite{ma2023glyphdraw} pioneered using glyph conditions and location masks for visual text generation of complex characters. It employs a glyph image and a location mask to control content and placement, though it remains limited to one line per image. TextDiffuser~\cite{chen2024textdiffuser} introduces a dedicated layout generation module to produce character-level masks as diffusion model conditions, enhancing Latin text generation. But it still cannot generate non-Latin texts. TextDiffuser-2~\cite{DBLP:conf/eccv/ChenHLCCW24} refines this approach by replacing character-level masks with bounding box coordinates, but this modification weakens its capacity for flexible visual text layout customization. GlyphControl~\cite{yang2024glyphcontrol} leverages ControlNet~\cite{zhang2023adding} and uses rendered glyph images as control conditions, enabling visual text generation in models like Stable Diffusion while preserving their core image synthesis capabilities. Nevertheless, GlyphControl is restricted to straight-line text layouts and frequently generates extraneous text artifacts. Building on these works, AnyText~\cite{tuo2023anytext} proposes a unified framework for high-quality multilingual visual text generation. Its text perceptual loss directly evaluates text accuracy during training, improving correctness of generated visual text. However, AnyText struggles with rendering text rotated beyond 45 degrees. TextGen~\cite{zhang2024control} examines how control signals affect image generation across timesteps. TextHarmony~\cite{DBLP:conf/nips/ZhaoTWLWL00H024} unifies image comprehension, generation, and editing within a single framework by training a multimodal Large Language Model (LLM) to improve image understanding and generate more precise tokens for image synthesis. However, users face challenges in defining text layout during image generation, as current methods rely solely on textual prompts. In terms of text editing, its performance in complex backgrounds is unsatisfactory. Diff-Text leverages Canny ControlNet for glyph-based generation but ties text placement to predefined objects in prompts. It may fail if a prompt does not include predefined objects, leading to unnatural images. 
 
\section{Method}
\label{sec:mthd}

\begin{figure*}[htbp] 
	\centering
	\includegraphics[width=\linewidth,scale=1.00]{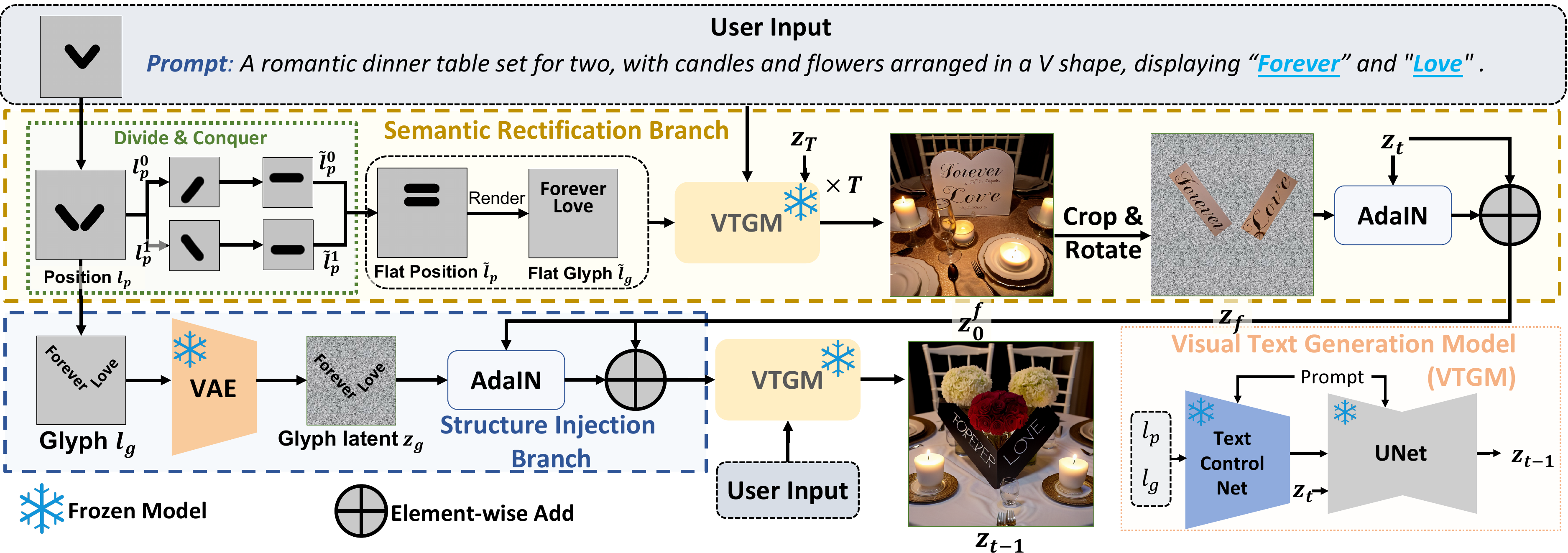}
    \vspace{-20pt}
	\caption{\textbf{Pipeline of our method}. \textit{Given user input on the leftmost side, which contains a prompt and a mask $l_p$ specifying positions for generating visual texts, we first split the $l_p$ using Divide and Conquer Strategy and obtain glyph image $l_g$ and flat position mask $\Tilde{l}_p$. Then $\Tilde{l}_p$ and $l_g$ are input to the Semantic Rectification Branch and Structure Injection Branch respectively. In the Semantic Rectification Branch, based on $\Tilde{l}_p$, we render flat glyph $\Tilde{l}_g$ and use them along with prompt and random noise $z_T$ to generate the latent with flat visual text. This latent serves as a semantic prior, providing rich semantic information for both the generation of the text and its background. $l_g$, on the other hand, is converted into the latent space as a structural prior for structural refinement of the text. Finally, the two prior combined to guide the generation of the visual text in $l_p$.}}
    \vspace{-10pt}
	\label{fig:pipeline}
\end{figure*}

Given a prompt $y$ describing the image and a challenging position mask $l_p$, 
our goal is to generate an image based on $y$ that incorporates visual text at the specified positions dictated by $l_p$.
Current visual text generation models struggle to generate visual texts in more challenging $l_p$ settings, such as tilted or curved text layouts. 

Our method aims to solve the problem based on two key insights: First, when the position mask $l_p$ is flat, current models can generate visual texts with high accuracy. We leverage this proficiency with flat visual texts to address challenging scenarios, providing a strong semantic prior for generation. Second, the glyph image contains little semantic information but is rich in glyph structural details. It can serve as a structure prior to further refine the structural information of the visual texts within the latent, thus enabling it to generate accurate slanted visual texts.

Our approach consists of two main branches: the \textit{Semantic Rectification Branch} and the \textit{Structure Injection Branch}. 
As shown in \cref{fig:pipeline}, the semantic rectification branch first takes the masks $\Tilde{l}_p$ reconfigured from $l_p$, prompt $y$, and random noise $z_T$ to generate the latent representation $z_0^f$, which contains reconfigured and flat visual text as a semantic prior. The structure injection branch then takes the rendered glyph image $l_g$ and $z_t$ to generate a structural prior. These two priors are merged and fed into the VTGM, along with $y$, $l_p$, and $l_g$, for the denoising process. Further details of these branches are discussed in \cref{sec:srb} and \cref{sec:sib}.


\subsection{Semantic Rectification Branch}
\label{sec:srb}



In the initial stages of the DDIM denoising process in the VGTM, 
clear text is generated. This text remains accurate when a flat position mask is applied. However, when using a slanted mask, the text gradually becomes distorted.
The reason for the distortion lies in the semantic drift in the text region when tilted. Motivated by this observation, we inherit the high-quality flat text generation capabilities of the existing visual text model, adopting the latent of flat text as a constant reference to rectify the semantic information for challenging text generation.



\paragraph{Reference Branch for Semantic Rectification.} Parallel to the generation branch, a separate branch is employed to generate flat visual texts using the same prompt $y$ for additional semantic information, which we refer to as the semantic rectification branch. In this branch, the flat visual text latent is blended into the tilted one within the text region, while the rest of the latent remains unchanged.

As shown in \cref{fig:pipeline}, the flat position $\Tilde{l}_p$ and corresponding glyph image $\Tilde{l}_g$ are fed into the VTGM along with prompt $y$ and random noise $z_T$. After $T$ denoising steps, we obtain the reference latent $z^f_0$. We rotate $z^f_0$ to match the user-given position, and extract its text region $z_f$ as a semantic guide for the branch below:
\begin{equation}
\label{eq:SemRB}
    \Tilde{z}_t = z_{f} \odot l_p + z_t \odot (1-l_p),
\end{equation}
where $l_p$ is the position mask input by the user. This operation effectively rectifies the visual text using the accurate text in the $z_f$. Thanks to the faithful background semantic information embedded in the $z_f$, we significantly reduce the erroneous non-textual semantic information in the text region while maintaining a coherent background, thereby avoiding background occlusion. 

\paragraph{AdaIN Combination.} AdaIN~\cite{huang2017arbitrary} is originally developed for style transfer tasks. It substitutes mean and standard deviation of the source feature with those of the target feature. 
Masui \etal~\cite{masui2024harnessing} demonstrated that AdaIN can be applied directly to Diffusion models for style transfer without any additional training. In our method, we also incorporate AdaIN, but to preserve image integrity against disruptions in latent distribution caused by the replacement operation:  
\begin{equation}
\begin{aligned}
    \Tilde{z}_t &= {\rm AdaIN}(z_f,z_t) \odot l_p + z_t \odot (1 - l_p) \\
    &= (\sigma(z_t)(\frac{x-\mu(z_f)}{\sigma(z_f)})+\mu(z_t))\odot l_p + z_t \odot (1- l_p),
\end{aligned}
\label{eq:AdaIN_glyph}
\end{equation}
where $\mu(\cdot)$ and $\sigma(\cdot)$ denote the channel-wise mean and standard deviation. The aid of AdaIN ensures the latent for guidance has the same range as the original latent, minimizing the change in latent distribution. With a more consistent latent, the visual text semantic information is further enhanced while preserving the overall image coherence. 
\paragraph{Divide and Conquer Strategy.} 
In practice, complex position masks are often provided for generating realistic visual texts, which often consist of multiple straight or curved parts, bringing obstacles to the geometric alignment to flat position masks $\Tilde{l}_p$ in our method. To deal with the problem, we propose a divide-and-conquer strategy to effectively segment complex position masks into more manageable straight sections as shown in \cref{fig:pipeline}. 
We use Bézier curves to define the upper and lower boundaries of $l_p$ and establish a baseline for the texts by averaging these curves. 
We identify splitting points on the curve where the direction vectors at each point are parallel to the boundaries of the minimum bounding box of $l_p$, therefore creating $N$ reconfigured masks:
\begin{equation}
    l_p = \left\{l^0_p, l^1_p, \ldots, l^{N-1}_p \right\}.
\end{equation}

Subsequently, we rotate and regroup these reconfigured masks to obtain flat position masks $\Tilde{l}_p$ and render corresponding glyphs $\Tilde{l}_g$:

\begin{equation}
\Tilde{l}_p =\left\{\Tilde{l}^0_p,\Tilde{l}^1_p,\ldots,\Tilde{l}^{N-1}_p\right\},
\Tilde{l}_g = \left\{\Tilde{l}^0_g,\Tilde{l}^1_g,\ldots,\Tilde{l}^{N-1}_g \right\}.
\end{equation}

With the assistance of this branch, 
we can generate highly accurate texts and create harmonious, diverse images by leveraging rich semantic information in both the visual texts and background.

\subsection{Structure Injection Branch}
\label{sec:sib}
Despite semantic guidance, the predictions may still deviate due to limited information in structure, causing cumulative errors and structural inconsistencies that require enriched structural guidance for clarity.
To resolve this, we introduce a glyph structure prior directly into the latent space. Unlike early methods~\cite{ma2023glyphdraw} that use glyph images directly as auxiliary information, we consider text glyphs as essential parts of the image, which should be comprehended by the visual text generation model in the latent space. As shown in \cref{fig:pipeline}, the structure injection branch feeds the rendered glyph image $l_g$ to Variational Autoencoder (VAE)~\cite{DBLP:journals/corr/KingmaW13} encoder to obtain the latent $z_g$. As an extension of semantic prior, we incorporate the $z_g$ as a structure prior into the latent $z_t$, using the same AdaIN operation above to regulate the range, which is represented as:
\begin{equation}
\label{eq:add_glyph}
    \hat{z}_t = {\rm AdaIN}(z_g,z_t) \odot l_p + z_t \odot (1-l_p).
\end{equation}
This operation provides the model with a structural foundation that serves as a strong starting point in the generation process, further enhancing the overall structure of the visual texts. 

Combining the two branches, the merged prior is represented by:
\begin{equation}
\label{eq:SPB}
    \hat{z}_t = \rho {\rm AdaIN}(z_g,z_t) + (1 - \rho) \Tilde{z}_t,
\end{equation}
where $\rho$ is a hyper-parameter that adjusts the balance between semantic and structural information. The modified latent is represented as follows:
\begin{equation}
    \ddot{z}_t = (\kappa_t\lambda \hat{z}_t + (1 - \kappa_t) z_t) \odot l_p + z_t \odot (1-l_p),
\end{equation}
where $\lambda$ is a hyper-parameter and the $\kappa_t$ is a temporal factor that decays over time with each timestep. Together, they control the injection strength of the merged prior.

\section{Experiments}
\label{sec:exper}

\begin{table}[t]
\footnotesize
  \centering
  \begin{tabular}{cccccc}
    \toprule
    Language & $\lambda$ & $\rho$ & Sen.Acc$\uparrow$ & NED$\uparrow$ & CLIP Score$\uparrow$ \\
    \midrule
    \multirow{6}{*}{English} & -0.5 & \cellcolor{gray!30}{0.5} & 44.88 & 64.11 & 0.3005  \\
    ~                        & 0.5 & \cellcolor{gray!30}{0.5}  & \textbf{45.43} & 65.10 & \underline{0.3027} \\
    ~                        & \cellcolor{gray!30}{0.5} & 0.25 & \underline{45.12} & 63.85 & \textbf{0.3036}  \\
    ~                        & \cellcolor{gray!30}{0.5} & 0.75 & 45.10 & \textbf{65.65} & 0.3006  \\
    ~                        & \cellcolor{gray!30}{0.5} & 1.50 & 45.01 & \underline{65.34} & 0.3009 \\
    ~                        & \cellcolor{gray!30}{0.5} & 2.00 & 44.53 & 65.17 &  0.3009 \\
    \midrule
    \multirow{6}{*}{Chinese} & -0.5 & \cellcolor{gray!30}{0.5}  & 49.28 & 87.29 & 0.3067  \\
    ~                        &  0.5 & \cellcolor{gray!30}{0.5}  &  \underline{49.96} & \textbf{88.08} & \underline{0.3071} \\
    ~                        & \cellcolor{gray!30}{0.5} & 0.25 & \textbf{50.70} & \underline{87.86} & \textbf{0.3076}  \\
    ~                        & \cellcolor{gray!30}{0.5} & 0.75 & 49.05 & 87.89 & 0.3058  \\
    ~                        & \cellcolor{gray!30}{0.5} & 1.50 & 47.40 & 87.64 & 0.3061  \\
    ~                        & \cellcolor{gray!30}{0.5} & 2.00 & 47.88 & 87.63 &  0.3059 \\
    \bottomrule
  \end{tabular}
  \vspace{-5pt}
  \caption{\textbf{Sensitivity analysis for $[\lambda, \rho]$.}}
  \vspace{-15pt}
  \label{tab:hyper}
\end{table}

\begin{table*}[ht]
\small
  \centering
  \begin{tabular}{ccccccccccc}
    \toprule
    \multirow{2}*{Language} & \multirow{2}*{Methods}  & \multicolumn{4}{c}{Sen.Acc$\uparrow$} & \multicolumn{4}{c}{NED$\uparrow$} & \multirow{2}*{CLIP Score$\uparrow$} \\
    \cmidrule(l){3-6} \cmidrule(l){7-10} 
    ~ & ~ & easy & medium & hard & total & easy & medium  & hard & total & ~\\
    \midrule
    \multirow{10}{*}{\rotatebox{90}{English}} & TextDiffuser~\cite{chen2024textdiffuser}   & 49.88 & 20.10 & 0.262 & 29.06 & 70.86 & 41.25 & 4.47 & 45.78 & 0.3091 \\
    ~ & TextDiffuser-2~\cite{DBLP:conf/eccv/ChenHLCCW24}   & 0.59 & 0.00 & 0.15 & 0.32 & 3.23 & 0.60 & 0.50 & 1.81 & 0.2989 \\
    ~ & SD1.5+TextHarmony$^{\dag}$~\cite{DBLP:conf/nips/ZhaoTWLWL00H024}   & 1.14 & 0.00 & 0.15 & 0.58 & 16.85 & 2.51 & 1.12 & 8.92 & 0.3090 \\
    ~ & GlyphControl~\cite{yang2024glyphcontrol}  & 19.00 & 1.63 & 0.22 & 9.47 & 41.74 & 9.17 & 2.98 & 22.95 & \textbf{0.3206} \\
    ~ & SD1.5+SceneVTG$^{\dag}$~\cite{zhu2024visualtextgenerationwild}  & 13.62 & 4.97 & 1.12 & 8.06 & 24.14& 13.40 & 3.00 & 15.80 & 0.3112 \\
    ~ & Anytext~\cite{tuo2023anytext}  & \underline{62.77} & 29.12 & 2.02 & 38.04 & \underline{83.64} & \underline{56.99} & 14.23 & 58.59 & 0.3007 \\
    \cmidrule(l){2-11}
    ~ & Diff-Text~\cite{zhang2024brush}   & 40.13 & 26.42 & \underline{8.85} & 28.38 & 61.44 & 45.37 & 17.94 & 45.91 & 0.2962\\
     ~ & Diff-Text+Ours  & 62.23 & \textbf{50.99} & \textbf{27.28} & \textbf{50.21} & 76.49 & 64.34 & \textbf{37.49} & \underline{63.18} & 0.3018\\
    ~ & GlyphControl+Ours  & 39.19 & 2.63 & 1.05 & 19.48 & 55.57 & 11.83 & 6.74 & 31.17 & \underline{0.3175} \\
    ~ & AnyText+Ours  & \textbf{71.25} & \underline{37.25} & 6.60 & \underline{45.43} & \textbf{87.54} & \textbf{64.42} & \underline{24.56} & \textbf{65.10} & 0.3027 \\
    \midrule
    \multirow{10}{*}{\rotatebox{90}{Chinese}} & TextDiffuser~\cite{chen2024textdiffuser}   & 5.41 & 4.07 & 0.09 & 4.07 & 56.81 & 49.37 & 43.01 & 52.47 & 0.3066 \\
    ~ & TextDiffuser-2~\cite{DBLP:conf/eccv/ChenHLCCW24}   & 0.06 & 0.08 & 0.00 & 0.05 & 7.28 & 4.81 & 6.04 & 6.49 & 0.2984 \\
    ~ & SD1.5+TextHarmony$^{\dag}$~\cite{DBLP:conf/nips/ZhaoTWLWL00H024}   & 0.00 & 0.00 & 0.00 & 0.00 & 14.35 & 8.50 & 12.00 & 12.60 & 0.3104 \\
    ~ & GlyphControl~\cite{yang2024glyphcontrol} & 2.33 & 0.24 & 0.0 & 1.41 & 50.98 & 15.58 & 35.94 & 40.24 & \textbf{0.3192} \\
    ~ & SD1.5+SceneVTG$^{\dag}$~\cite{zhu2024visualtextgenerationwild} & 1.60 & 0.81 & 0.18 & 1.14 & 7.33 & 4.62 & 4.61 & 6.20 & 0.3025\\
    ~ & Anytext~\cite{tuo2023anytext} & \underline{66.00} & 26.38 & 2.02 & \underline{44.78} & \underline{94.48} & \underline{81.37} & 59.38 & \underline{84.74} & 0.3064 \\
    \cmidrule(l){2-11}
    ~ & Diff-Text~\cite{zhang2024brush} & 23.46 & 17.26 & 7.35 & 18.95 & 68.27 & 62.71 & 48.65 & 63.21 & 0.2950\\
    ~ & Diff-Text+Ours & 42.13 & \textbf{38.43} & \textbf{22.42} & 37.47 & 81.15 & 71.84 & \underline{59.69} & 74.91 & 0.2978\\
    ~ & GlyphControl+Ours & 6.27 & 0.57 & 0.74 & 3.93 & 62.50 & 18.08 & 45.58 & 49.40 & \underline{0.3168} \\
    ~ & AnyText+Ours  & \textbf{69.22} & \underline{35.58} & \underline{8.55} & \textbf{49.96} & \textbf{95.37} & \textbf{85.85} & \textbf{68.77} & \textbf{88.08} & 0.3071 \\
    \bottomrule
  \end{tabular}
  \vspace{-5pt}
  \caption{\textbf{Quantitative Comparison between STGen and other competitors on both English and Chinese sets. } \textit{All competitors are evaluated based on their officially released code and models. Numbers in \textbf{bold} indicate the best performance, and \underline{underscored} numbers indicate the second best. $^\dag$ indicates we adapt the method which is originally focused on other tasks into image generation.}}
  \vspace{-8pt}
  \label{tab:exper}
\end{table*}

\begin{table}[ht]
\small
    \centering
    \begin{tabular}{lccc}
    \toprule
       Method & SEN.ACC $\uparrow$ & NED $\uparrow$ & FID $\downarrow$  \\
       \midrule
       GlyphControl & 37.10/3.27 & 66.80/8.45 & 37.84/34.36 \\
       w/ STGen & \textbf{70.92/11.81} & \textbf{84.68/24.72} & \textbf{32.00/32.99} \\
       \midrule
       Diff-Text & 56.11/29.49 & 75.28/48.67 & 69.24/62.61 \\
       w/ STGen &\textbf{68.55/39.58} &\textbf{81.55/55.61} & \textbf{66.77/61.59} \\
       \midrule
       AnyText & 72.39/69.23 & 87.67/83.96 & \textbf{33.54}/31.58 \\
       w/ STGen & \textbf{75.53}/\textbf{70.00} & \textbf{89.38}/\textbf{84.57} & 33.83/\textbf{32.17}  \\
       \bottomrule
    \end{tabular}
    \vspace{-5pt}
    \caption{\textbf{Evaluation on vanilla AnyText Benchmark.} \textit{In each cell of the table, the numbers on the right are results from the English set while the numbers on the right are from the Chinese set.}}
    \vspace{-8pt}
    \label{tab:anytextbench}
\end{table}

\subsection{Implementation Details}
\label{sec:imple}

We use a single RTX 3090 GPU for images of size (512, 512). Optimal performance is achieved when $\lambda \in [-0.5, 0.5]$ and $\rho \in (0, 2]$. We set $\kappa_t=10^{t-T}$ for effective guidance and harmonious text region boundary.
As shown in \cref{tab:hyper}, our method demonstrates robustness for hyper-parameter sensitivity. In subsequent experiments, we set $\lambda$ to $0.5$ and $\rho$ to $0.5$, which has a balanced performance on both Chinese and English for evaluation.

\subsection{Evaluation Setup}

Due to the lack of publicly available datasets focused on challenging visual text generation, we propose a new benchmark derived from the AnyText-benchmark~\cite{tuo2023anytext}. For each text position mask, we randomly rotate the original benchmark’s masks and resolve overlaps, yielding 984 prompts for LAION-word (English evaluation) and 919 prompts for Wukong-word (Chinese evaluation). 
During evaluation, masks are categorized into three difficulty levels based on rotation angles: easy (0°–30°), medium (30°–60°), and hard (60°–90°), enabling multi-level assessment of performance. 

Textual accuracy and background-text coherence are two main factors that determine the quality of slanted text generation, which we quantitatively evaluate through OCR accuracy. Following AnyText~\cite{tuo2023anytext}, 
we select the following two metrics for comparing OCR accuracy at word-level and character-level, respectively: (1) Sentence Accuracy (Sen.Acc); (2) Normalized Edit Distance (NED). 


We evaluated existing competing methods, including TextDiffuser~\cite{chen2024textdiffuser}, TextDiffuser-2~\cite{DBLP:conf/eccv/ChenHLCCW24}, TextHarmony~\cite{DBLP:conf/nips/ZhaoTWLWL00H024}, GlyphControl~\cite{yang2024glyphcontrol}, SceneVTG~\cite{zhu2024visualtextgenerationwild}, AnyText~\cite{tuo2023anytext} and Diff-Text~\cite{zhang2024brush} using the benchmark and metrics mentioned above. Notably, SceneVTG~\cite{zhu2024visualtextgenerationwild} cannot generate images from scratch. To address this, we use background images produced by Stable Diffusion~\cite{rombach2022high} with identical prompts. For TextHarmony~\cite{DBLP:conf/nips/ZhaoTWLWL00H024}, its image generation model lacks support for positioning visual text at specific locations. Following its evaluation protocol on the AnyText Benchmark, we employ its visual text editing mode, which generates text on images where target positions are masked in black. For all the baselines in the evaluation, we use their officially released code and checkpoints. To ensure fairness, we preprocess the masks using the \textit{Divide and Conquer} strategy mentioned above before generating images.

\begin{table}[ht]
\small
  \centering
  \begin{tabular}{lcc}
    \toprule
    Baseline & Baseline Preference & Ours Preference \\
    \midrule
    TextDiffuser~\cite{chen2024textdiffuser} & 23.07\% & 76.93\% \\
    TextDiffuser-2~\cite{DBLP:conf/eccv/ChenHLCCW24} & 20.15\% & 79.85\% \\
    GlyphControl~\cite{yang2024glyphcontrol} & 21.96\% & 78.04\% \\
    SceneVTG~\cite{zhu2024visualtextgenerationwild} & 18.99\% & 81.01\% \\
    TextHarmony~\cite{DBLP:conf/nips/ZhaoTWLWL00H024} & 18.75\% & 81.25\% \\
    AnyText~\cite{tuo2023anytext} & 25.31\% & 74.69\% \\
    Diff-Text~\cite{zhang2024brush} & 17.00\% & 83.00\% \\
    \bottomrule
  \end{tabular}
  \vspace{-5pt}
  \caption{\textbf{User study results.} \textit{Participants were asked to choose the best results based on image quality, accuracy of generated text within images, and prompt-image similarity. }}
  \vspace{-10pt}
  \label{tab:user study}
\end{table}

\begin{figure*}[ht]
    \centering
    \includegraphics[width=1.0\linewidth]{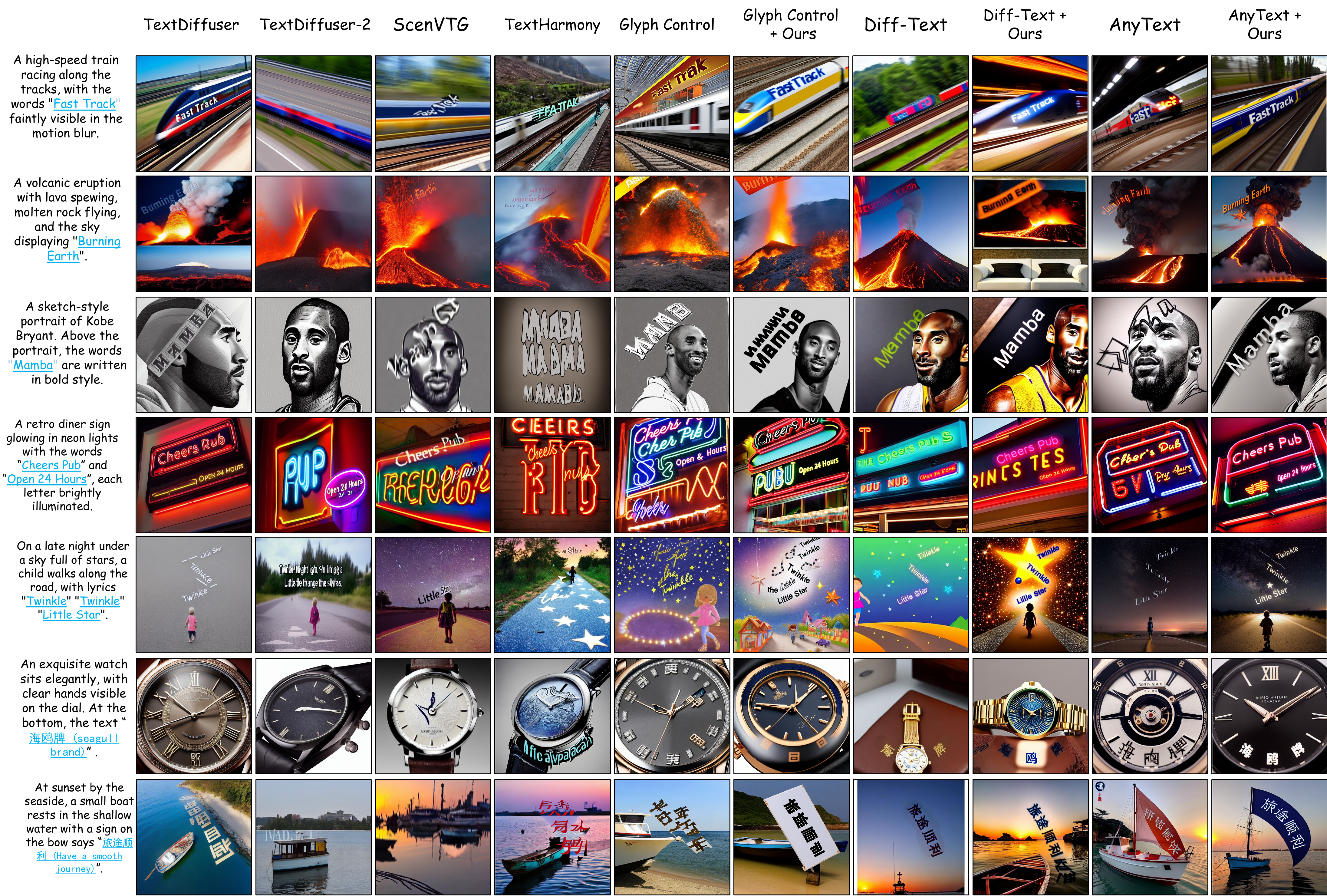}
    \vspace{-18pt}
    \caption{\textbf{Qualitative comparison of our method and state-of-the-art models in both English and Chinese text generation.}
    }
    \vspace{-8pt}
    \label{fig:qualitative}
\end{figure*}

\subsection{Quantitative Analysis}

\paragraph{OCR Accuracy.} 
As shown in \cref{tab:exper}, our method shows substantial improvements over the competitive methods in both English and Chinese across all levels. Notably, our method doesn't require additional training and enhances the performance of Diff-Text, GlyphControl, and AnyText at the easy level. At the hard level, AnyText with our method yields results closest to the ground truth in the NED metric, with an improvement of approximately $10\%.$ over the baseline AnyText in both languages. As similarly shown in \cref{tab:anytextbench}, our methods outperform baselines on vanilla AnyText Benchmark except FID. Meanwhile, Diff-Text with our method shows an impressive $20\%$ improvement at the hard level in the English set.

\paragraph{Text-Image Similarities.} In the absence of ground-truth images, we use the CLIP score~\cite{hessel2021clipscore} to assess the consistency between the prompt and the generated image. We compute the average cosine similarity between the prompt and the generated image, excluding the influence of the visual texts. As shown in \cref{tab:exper}, our method increases the accuracy of visual text generation without compromising the baseline performance. Additionally, our method slightly improves text-image similarity due to our design, which minimizes conflicts between the image composition and the visual text. This leads to a more coherent layout for both the image and the text.

\paragraph{User Study.} Following~\cite{bar2022text2live}, we conduct a user study to comprehensively compare the generation results in \cref{tab:user study}. We create 27 input sets of varying difficulty. For each set, participants receive our input prompt, position mask, and two images (our result and a baseline) in random order, and select the image with the best quality and most accurate visual text. The final score is the average number of selections per prompt. We gather 168 judgments from a diverse group of experts and non-experts and report vote percentages. As shown in \cref{tab:user study}, our method is preferred in all cases.

\begin{figure*}[ht]
    \centering
    \includegraphics[width=1.0\linewidth]{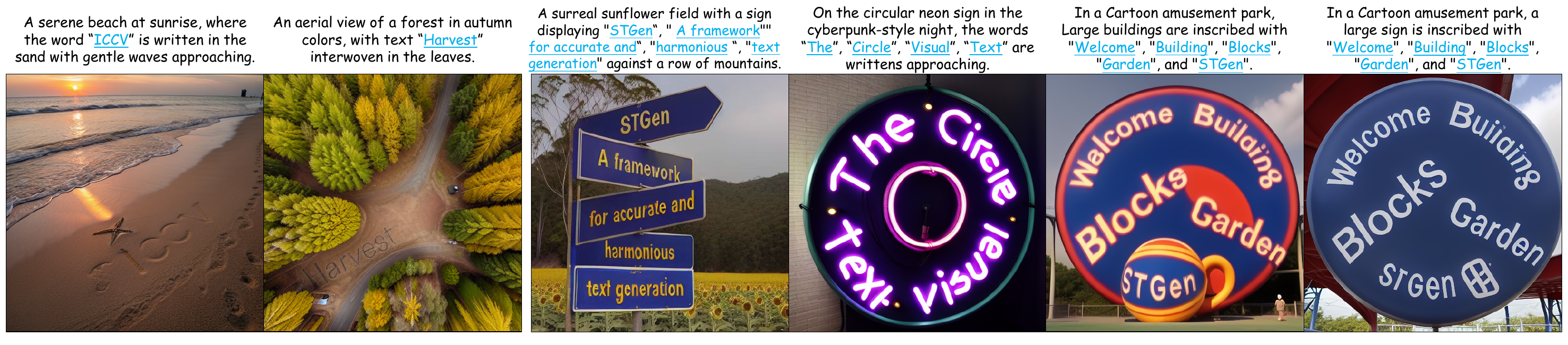}
    \vspace{-18pt}
    \caption{\textbf{More results on Background coherence and complex text layouts.} \textit{The left two columns show how our approach seamlessly blends text with the background, while the remaining images highlight its ability to produce multi-sentence and circular text. } 
    }
    \vspace{-12pt}
    \label{fig:bc}
\end{figure*}

\subsection{Qualitative Comparisons}

As presented in \cref{fig:qualitative}, TextDiffuser~\cite{chen2024textdiffuser} tends to generate low-quality images, specifically in the second and the fifth rows. Textdiffuser-2~\cite{DBLP:conf/eccv/ChenHLCCW24} fails to consistently generate legible text within specified bounding boxes. We suspect this occurs due to interference between textual and non-textual elements when using overly large bounding boxes. GlyphControl~\cite{yang2024glyphcontrol} frequently produces texts outside its designated area, particularly when the text is near the image boundary, as shown in the second and fourth row. 
SceneVTG~\cite{zhu2024visualtextgenerationwild} suffers from severe text distortion, especially in the first and third rows. Background occlusion is also evident in the second, third, and fourth rows, where the texts “Burning,” “Mamba,” and “Twinkle” are obscured by lava, a human figure, and stars, respectively. This suggests SceneVTG struggles with complex backgrounds and lacks seamless integration with other text-to-image models. TextHarmony~\cite{DBLP:conf/nips/ZhaoTWLWL00H024} faces similar issues. In image editing mode, it struggles to properly fill blanks with complex shapes, leading to distorted text, as seen in the second row. For Diff-text~\cite{zhang2024brush}, there is a noticeable lack of coherence in the images due to insufficient consideration for harmony between the visual texts and the background. 
AnyText~\cite{tuo2023anytext} tends to produce distorted visual texts due to a loss of semantic and structural information during inference. For example, in the first row, the text “Track” overlaps with the train head, compromising both elements. In contrast, our method generates high-quality images with accurately rendered English and Chinese texts—even in challenging scenarios such as those in the fourth and fifth rows and \cref{fig:bc}. Moreover, our approach not only corrects text distortion but also reduces background occlusion between textual and non-textual elements. For instance, in the third row of \cref{fig:qualitative}, AnyText~\cite{tuo2023anytext} disrupts the text “Mamba” with a human head, whereas our method repositions the head to produce clear text alongside well-integrated non-text elements. 



\begin{figure}[t]
    \centering
    \includegraphics[width=1.0\linewidth]{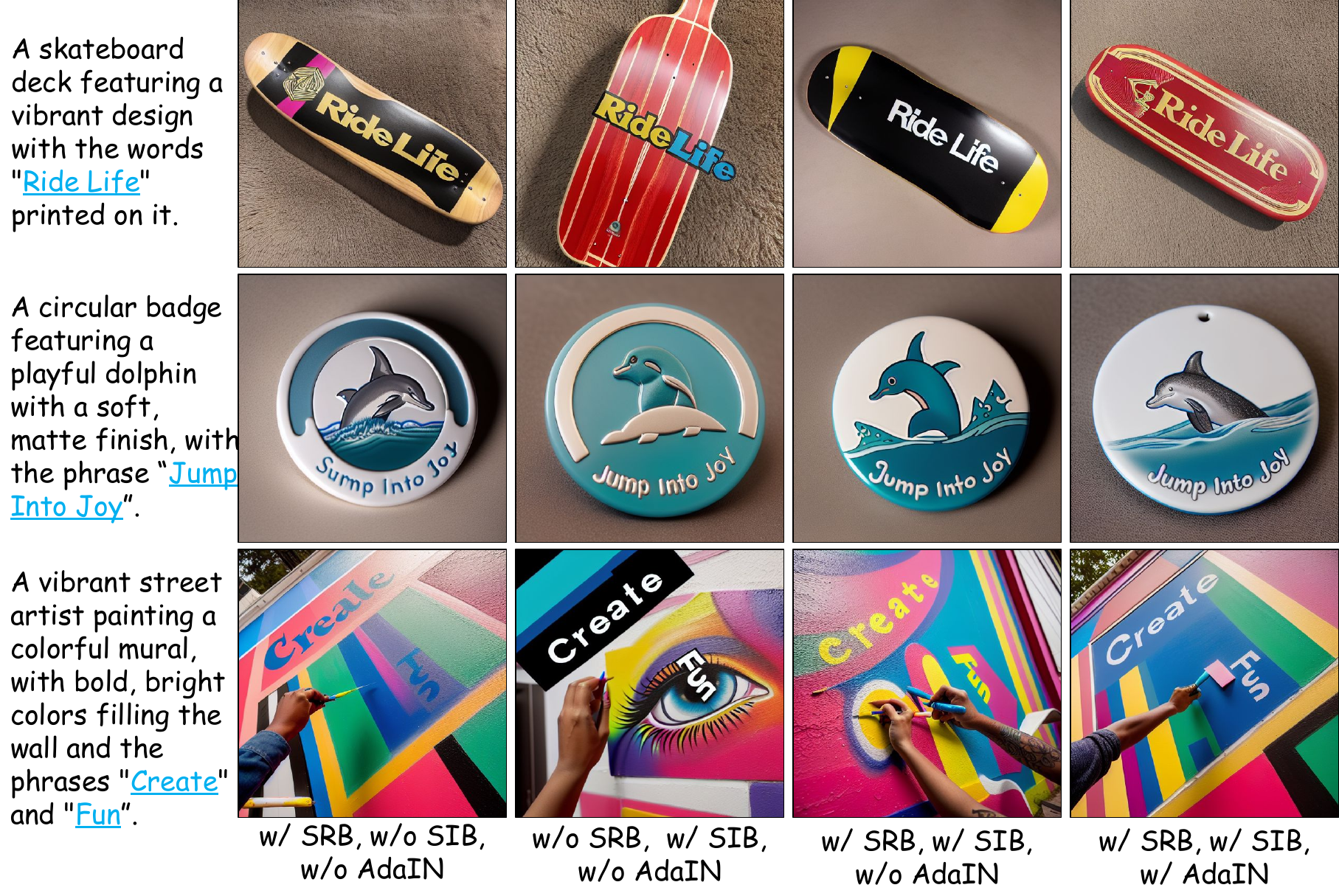}
    \vspace{-15pt}
    \caption{\textbf{Ablation Study Visualization.} \textit{The first column shows accuracy improvement with SRB. The second reveals that SIB enhances accuracy but compromises coherence. The third confirms that combining both branches enhances both. Finally, the fourth shows that adding AdaIN further improves image quality.}
    }
    \vspace{-15pt}
    \label{fig:ablation}
\end{figure}

\section{Ablation Study}
\label{sec:ablation}

In this section, we demonstrate the effectiveness of our components through an ablation study on the medium level. We use OCR accuracy (Sen.Acc and NED) and CLIP score as the main metrics, given their importance in evaluating visual text generation. All parameters are kept consistent with those outlined in \cref{sec:imple}. 

\paragraph{SRB without AdaIN.} As shown in the first two rows of \cref{tab:ablation}, the addition of \textit{Semantic Rectification Branch} improves the visual text accuracy. Similarly, as demonstrated by the first column of \cref{fig:ablation}, 
the generated visual text has a clear structure. However, this also negatively impacts text-image consistency, as indicated by the slight drop in the CLIP score. This occurs because the operation of replacing partial latent disrupts the latent distribution and affects the representation of other parts of the image. Thanks to the rich semantic information in the latent, the drop is minor. 

\begin{table}[t]
  \centering
  \footnotesize
  \begin{tabular}{cccccc}
    \toprule
     SRB & SIB &  AdaIN & Sen. Acc & NED & CLIP Score \\
    \midrule
     \ding{55} & \ding{55} & \ding{55} &  29.12 & 56.99 & 0.3007 \\
     \ding{51} & \ding{55} & \ding{55} & 36.22 & 61.17 & 0.3003 \\ 
     \ding{51} & \ding{51} & \ding{55} & 36.93 & 64.35 & 0.3006 \\
     \ding{51} & \ding{51} & \ding{51}  & \textbf{37.25} & \textbf{64.42} & \textbf{0.3027} \\
    \bottomrule
  \end{tabular}
    \vspace{-5pt}
  \caption{\textbf{Ablation study.}}
  \vspace{-8pt}
  \label{tab:ablation}
\end{table}

\paragraph{SIB without AdaIN.} As illustrated in the second and third rows of \cref{tab:ablation}, \textit{Structure Injection Branch} further improves the accuracy. As similarly shown in 
the third column of \cref{fig:ablation}, this branch can structurally improve the text structure, resulting in impressive improvement. However, when applied alone, it may disrupt image coherence, as shown in the second column and third row of \cref{fig:ablation}, where the word 'Fun' occludes the eye. This issue arises from the lack of semantic information in the structural prior.

\paragraph{AdaIN Combination.} As revealed by the third and fourth column of \cref{fig:ablation}, AdaIN improves the accuracy and the harmony of the text and its background, lifting the CLIP score to a new level in the \cref{tab:ablation}. 


\section{Conclusion}
\label{sec:conclusion}

We advance visual text generation by tackling the challenge of complex text synthesis. Our proposed STGen introduces a dual-branch approach: the Semantic Rectification Branch, which refines text generation using latent extracted from simpler scenarios, and the Structure Injection Branch, which enhances text structure by incorporating latent of glyph image. For highly challenging cases, we break them into more manageable cases. Integrated via a dedicated ControlNet, STGen seamlessly enhances existing models. Extensive experiments on our benchmark confirm its superior performance, making STGen a promising step toward real-world applications.

{
    \small
    \bibliographystyle{ieeenat_fullname}
    \bibliography{main}
}
%

\end{document}